\RequirePackage{fix-cm}
\documentclass[twocolumn]{svjour3}          % twocolumn
\smartqed  % flush right qed marks, e.g. at end of proof

\usepackage{cite}
\usepackage[pdftex]{graphicx}
\usepackage{ragged2e}
\usepackage[tight,footnotesize]{subfigure}
\usepackage{rotating}
\usepackage{graphicx}
\usepackage{amsmath,amssymb,amsfonts} % define this before the line numbering.
\usepackage{array}
\usepackage{multirow}
\usepackage{colortbl}
\usepackage{amsfonts}
\usepackage{pifont}
\usepackage{xspace}
\usepackage{etoolbox}
\usepackage{overpic}
\usepackage{color}
\usepackage{microtype}
\usepackage{scalerel} 
\DeclareRobustCommand\fire{\scalerel*{\includegraphics{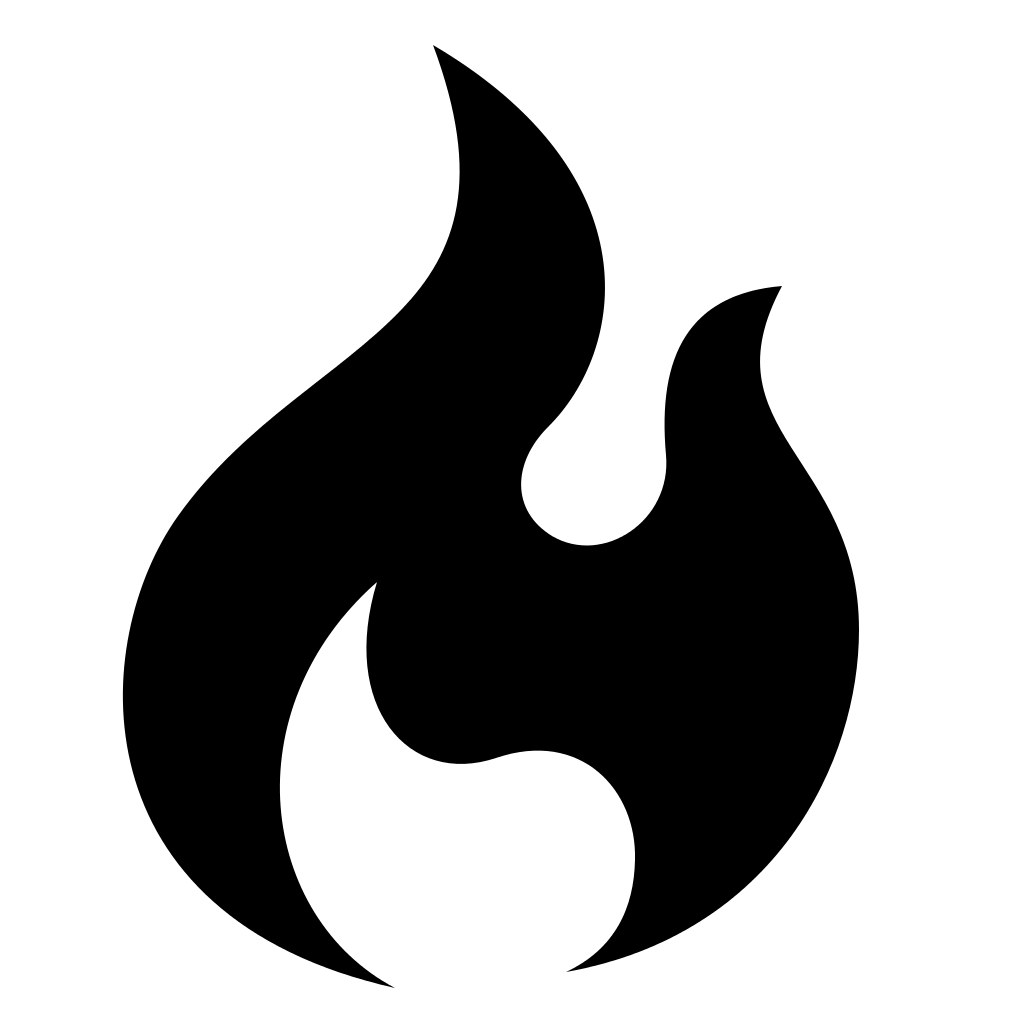}}{X}}
\usepackage{anyfontsize}
\usepackage[hyphens]{url}   % allow breaks at hyphens
\usepackage{xurl}

\newcommand{\orcid}[1]{\href{https://orcid.org/#1}{\includegraphics[width=10pt]{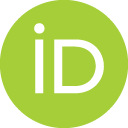}}}

\def\etal{{\em et al}}

\usepackage{hyperref}
\hypersetup{breaklinks=true,citecolor=blue, colorlinks}

\DeclareGraphicsExtensions{.pdf,.jpg,.png}
\usepackage{booktabs}%

\usepackage{silence}
\journalname{Research Article}

\begin{document}

\title{When SAM2 Meets Video Camouflaged Object Segmentation: A Comprehensive Evaluation and Adaptation}

% \titlerunning{Short form of title}        % For running head

\author{Yuli Zhou$^{1,2}$ \orcid{0000-0003-4584-1893}        \and
  Guolei Sun$^{1*}$ \orcid{0000-0001-8667-9656} \and 
  Yawei Li$^{1,3}$ \orcid{0000-0002-8948-7892} \and
  Guo-Sen Xie$^{4}$ \orcid{0000-0002-5487-9845} \and
  Luca Benini$^{3,5}$ \orcid{0000-0001-8068-3806} \and
  Ender Konukoglu$^{1}$ \orcid{0000-0002-2542-3611}
}

\authorrunning{Y. Zhou \etal} % if too long for running head

\institute{
Yuli Zhou, Guolei Sun, Yawei Li, and Ender Konukoglu are with Computer Vision Laboratory, ETH Z\"urich, Z\"urich, Switzerland. 
(Email: yulzhou@vision.ee.ethz.ch, guolei.sun@vision.ee.ethz.ch, yawei.li@vision.ee.ethz.ch, ender.konukoglu@vision.ee.ethz.ch). \\
% Yuli Zhou is with the University of Zurich, Zurich, Switzerland. \\
Yawei Li and Luca Benini are with Integrated System Laboratory, ETH Z\"urich, Z\"urich, Switzerland. 
(Email: yawei.li@vision.ee.ethz.ch, lbenini@iis.ee.ethz.ch). \\
Guo-Sen Xie is with the School of Computer Science and Engineering, Nanjing University of Science and Technology, Nanjing, China.
(Email: gsxiehm@gmail.com). \\
% Luca Benini is with the University of Bologna, Bologna, Italy. \\
Corresponding author: Guolei Sun \\ (guolei.sun@vision.ee.ethz.ch).
}

\date{Received: date / Accepted: date}

\maketitle

\begin{abstract}
This study investigates the application and performance of the Segment Anything Model 2 (SAM2) in the challenging task of video camouflaged object segmentation (VCOS). VCOS involves detecting objects that blend seamlessly in the surroundings for videos due to similar colors and textures and poor light conditions. Compared to the objects in normal scenes, camouflaged objects are much more difficult to detect. SAM2, a video foundation model, has shown potential in various tasks. However, its effectiveness in dynamic camouflaged scenarios remains under-explored. This study presents a comprehensive study on SAM2's ability in VCOS. First, we assess SAM2's performance on camouflaged video datasets using different models and prompts (click, box, and mask). Second, we explore the integration of SAM2 with existing multimodal large language models (MLLMs) and VCOS methods. Third, we specifically adapt SAM2 by fine-tuning it on the video camouflaged dataset. Our comprehensive experiments demonstrate that SAM2 has the excellent zero-shot ability to detect camouflaged objects in videos. We also show that this ability could be further improved by specifically adjusting SAM2's parameters for VCOS. 
% The code is available at \url{https://github.com/zhoustan/SAM2-VCOS}.

% Please provide 4 to 6 keywords which can be used for indexing purposes.

\end{abstract}

\section{Introduction}
\label{sec:intr}
Camouflaged object segmentation, aiming to identify objects that blend into their surroundings, is a fundamental task in computer vision. Unlike conventional segmentation tasks, where objects typically exhibit distinct boundaries, camouflaged objects often have similar colors or textures to the background, making them difficult to be perceived. This task becomes even more complex for video sequences, where both objects and background can change dynamically over time.

Traditional segmentation or detection approaches~\cite{long2015fully,he2017mask,huang2024m,sun2024learning}, such as fully convolutional networks (FCNs)~\cite{long2015fully} and mask R-CNN~\cite{he2017mask}, have made significant contributions to the field of object detection and segmentation in common scenes where objects are generally distinguishable from the environment. However, these models often struggle with camouflaged scenes. More recent techniques such as SINet~\cite{fan2020Camouflage}, SLT-Net~\cite{cheng2022implicit}, and ZoomNeXt~\cite{pang2024zoomnext} have been specifically designed to address the challenges of camouflaged object detection. 
Despite these advancements, existing video camouflaged object segmentation (VCOS) methods still face significant challenges in maintaining temporal consistency, as they often lack effective object-level memory, making it difficult to achieve stable segmentation across long video sequences. Some approaches, such as Flow-SAM~\cite{xie2024movingobjectsegmentationneed} and SLT-Net~\cite{cheng2022implicit}, incorporate flow-guided or motion-guided networks to enhance temporal coherence. However, these methods struggle when camouflaged objects remain stationary or exhibit minimal movement, leading to segmentation failures. Additionally, they are prone to errors in dynamic backgrounds, where optical flow may incorrectly associate background motion with the target object, further compromising segmentation accuracy.

Recently, the advanced large foundation model Segment Anything Model 2 (SAM2)~\cite{ravi2024sam} has shown promise in handling complex video-based segmentation tasks. It adopts prompt-based segmentation, where specific user-defined prompts such as bounding boxes, clicks, or masks are used to guide the segmentation process. This allows for great flexibility in segmenting target objects, particularly in scenarios like camouflaged detection, where object boundaries are usually invisible. Based on Segment Anything Model (SAM)~\cite{kirillov2023segment}, SAM2 is equipped with a memory module containing: a memory encoder, which encodes the predicted mask and image embedding of the current frame; a memory bank, which stores information about the object and previous interactions; and memory attention, which enhances the current frame's image embedding using past memories. This approach allows SAM2 to track camouflaged objects more reliably across video sequences.

Given SAM2's strong generalization ability and its flexible prompting mechanisms, we systematically investigate its effectiveness in addressing VCOS through three key aspects: 1) zero-shot capability, by directly adapting SAM2 on two camouflaged datasets, MoCA-Mask~\cite{cheng2022implicit} and CAD~\cite{bideau2016s}; 2) integration with state-of-the-art methods, by leveraging multimodal large language models (MLLMs) and existing VCOS techniques with SAM2, exploring hybrid approaches for VCOS; and 3) adaptation through task-specific fine-tuning on the largest-scale VCOS benchmark MoCA-Mask. Our study is organized into three parts, with each targeting a specific aspect of this complex problem:

\noindent 1) \textbf{Evaluating SAM2's zero-shot ability for VCOS}. We thoroughly assess the performance of SAM2 in segmenting camouflaged objects, an inherently challenging task due to the high degree of similarity between the object and its background. SAM2 is evaluated in two modes: \textit{automatic} and \textit{semi-supervised} mode. In automatic mode, we leverage SAM2's built-in automatic mask generator to produce initial segmentation masks, which are then used as mask prompts for subsequent frames. In semi-supervised mode, we investigate how manual interactions (via click, box, and mask-based prompts) and prompting time can impact segmentation quality. This analysis provides the first detailed exploration of SAM2's behavior specifically for VCOS, a domain where traditional segmentation models typically struggle.

\noindent 2) \textbf{Augmenting SAM2 with state-of-the-art MLLMs and VCOS methods}. We further explore the effectiveness of SAM2 when combined with advanced MLLMs and VCOS techniques. For MLLMs, we design specific prompts to generate bounding boxes around potential camouflaged regions, which are then used by SAM2 as box prompts for segmentation. For VCOS, we enhance segmentation by refining the initial mask generated by VCOS methods with SAM2's powerful refinement capabilities. Our contribution here is the novel integration of SAM2 with MLLMs and VCOS methods, demonstrating how prompt-driven refinement can improve segmentation accuracy in highly complex visual scenes.
    
\noindent 3) \textbf{Fine-tuning SAM2 on largest-scale VCOS benchmark}. We explore how task-specific fine-tuning can improve SAM2's segmentation performance in VCOS. By fine-tuning SAM2 on the MoCA-Mask dataset, we extend SAM2's utility beyond its initial design, showing how it can be adapted to specialized datasets for enhanced performance.

Our contributions can be summarized as follows:

\noindent 1) We provide the first thorough evaluation and in-depth analysis of SAM2's performance in the challenging task of VCOS, experimenting with both automatic and semi-supervised modes.

\noindent 2) We propose a novel hybrid approach that combines SAM2 with existing MLLMs and VCOS methods, demonstrating significant improvements in segmentation accuracy through prompt-based refinements.

\noindent 3) We enhance SAM2's capabilities by adjusting its parameters on the well-known MoCA-Mask dataset to better fit VCOS, achieving state-of-the-art results.

\section{Related Work}
\subsection{Camouflaged Object Detection}
Camouflaged scene understanding (CSU) aims to perceive the scenes where objects are difficult to be distinguished with the background. These scenes commonly exist in natural environments, such as forests, oceans, and deserts. Among various CSU tasks~\cite{fan2020Camouflage,lv2021simultaneously,sun2023indiscernible,pei2022osformer,luo2024vscode}, camouflaged object detection (COD) attracts lots of research attention, which is to identify objects that blend seamlessly into their environment, posing a significant challenge for conventional perception techniques. Traditional approaches to object detection, which rely on strong edge features, color contrasts, and texture variations, often fail when applied to camouflaged objects. This led to the development of specialized COD models that incorporate unique feature extraction techniques, focusing on small differences in texture, edge irregularities, and subtle shifts in color or shading that indicate the presence of a hidden object~\cite{qin2021boundary, fan2020Camouflage}. Several innovative models have been proposed for COD. HGINet~\cite{yao2024hierarchical} uses a hierarchical graph interaction transformer with dynamic token clustering to capture multi-scale features for better object detection. AGLNet~\cite{chen2024adaptive} enhances COD performance by employing adaptive guidance learning, which adjusts feature extraction based on the object's appearance and context. PAD~\cite{xing2023pre} leverages a multi-task learning approach, pre-training on multiple datasets and fine-tuning for COD tasks through task-specific adaptation. DQNet~\cite{sun2022dqnet} focuses on cross-model detail querying, using multiple models to enhance the detection of subtle features in camouflaged objects. R2CNet~\cite{zhang2025referring} integrates linguistic cues with visual data to improve detection, particularly by using referring expressions to locate specific camouflaged objects. ZoomNeXt~\cite{pang2024zoomnext} presents a unified collaborative pyramid network to enhance feature extraction across multiple scales. WSSCOD~\cite{zhang2024learningcamouflagedobjectdetection} improves COD using weak supervision by learning from noisy pseudo labels, while CamoTeacher~\cite{lai2024camoteacher} applies dual-rotation consistency learning for semi-supervised detection, making use of limited labeled data. POCINet~\cite{liu2021integrating} integrates contrast information and part-object relational knowledge across search and identification stages, bridged by a search-to-identification guidance (SIG) module that enhances feature encoding through search results and semantic knowledge. STANet~\cite{liu2024camouflaged} introduces scale-feature attention and type-feature attention to enhance COD by integrating multi-scale information and contrast from CNNs and part-whole relationships from CapsNet. MCRNet~\cite{zhang2024mamba} is a Mamba capsule routing network, for part-whole relational COD, where a Mamba capsule generation (MCG) module enables lightweight type-level capsule routing, and a capsules spatial details retrieval (CSDR) module refines spatial details for final detection.
The development of these advanced COD models demonstrates significant progress in detecting camouflaged objects by integrating multi-scale feature extraction, adaptive learning, and novel supervision techniques, paving the way for more robust and accurate object detection in challenging environments.

\subsection{Video Camouflaged Object Segmentation}

Video camouflaged object segmentation (VCOS), an important task of CSU, aims to detect camouflaged objects in dynamic and video-based environments. To deal with VCOS, models are required to consider object motion, temporal continuity, and varying backgrounds, adding additional complexities to COD. Several advanced models have been introduced to tackle these challenges by integrating motion learning and temporal-spatial attention mechanisms. IMEX~\cite{hui2024implicit} introduces implicit-explicit motion learning, allowing for more robust detection of camouflaged objects through an integration of both implicit and explicit motion features. Temporal-spatial attention is also crucial for many VCOS models, as seen in TSP-SAM~\cite{hui2024endow}, which enhances SAM with a focus on temporal-spatial prompt learning to identify subtle object movements. SAM-PM~\cite{meeran2024sam} further extends this approach by applying spatio-temporal attention to boost accuracy in video sequences by tracking subtle changes in movement and background. An earlier method~\cite{cheng2022implicit} focuses on implicit motion handling to refine video object detection, particularly for scenes where motion is subtle or hard to detect. OCLR~\cite{lamdouar2023making} introduces three scores to automatically evaluate the effectiveness of camouflage by measuring background-foreground similarity and boundary visibility. These scores are used to enhance camouflaged datasets and integrate them into a generative model. MG~\cite{yang2021self} introduces a self-supervised Transformer-based model for motion segmentation using optical flow demonstrating the effectiveness of motion cues over visual appearance in VCOS. These models illustrate the importance of combining motion analysis with traditional object detection techniques to enhance VCOS performance in real-world applications.

\subsection{Segment Anything Model 2}
Segment Anything Model 2 (SAM2)~\cite{ravi2024sam} is a vision foundation model for segmenting objects across images and videos. SAM2 has shown excellent performances in medical video and 3D segmentation, including tasks such as polyp detection, surgical video segmentation, and other medical image segmentation~\cite{mansoori2024polypsam2advancing, shen2024performancenonadversarialrobustnesssegment, yu2024sam2roboticsurgery, liu2024surgical, he2024shortreviewevaluationsam2s, chen2024sam2adapterevaluatingadapting, xiong2024sam2unetsegment2makes}. Furthermore, SAM2 has been applied in the segmentation of 3D meshes and point clouds~\cite{tang2024segmentmeshzeroshotmesh}, remote sensing~\cite{rafaeli2024promptbasedsegmentationmultipleresolutions}, and image camouflaged object detection~\cite{chen2024sam2adapterevaluatingadapting, xiong2024sam2unetsegment2makes, tang2024evaluatingsam2srolecamouflaged}.

Despite these advances, to the best of our knowledge, there is no specific study focusing on VCOS using SAM2. This study fills this gap by systematically examining SAM2's performance in VCOS and proposing strategies to further improve SAM2's capability.

\section{Methods}

\begin{figure*}[!t]
    \centering
    \includegraphics[width=\linewidth]{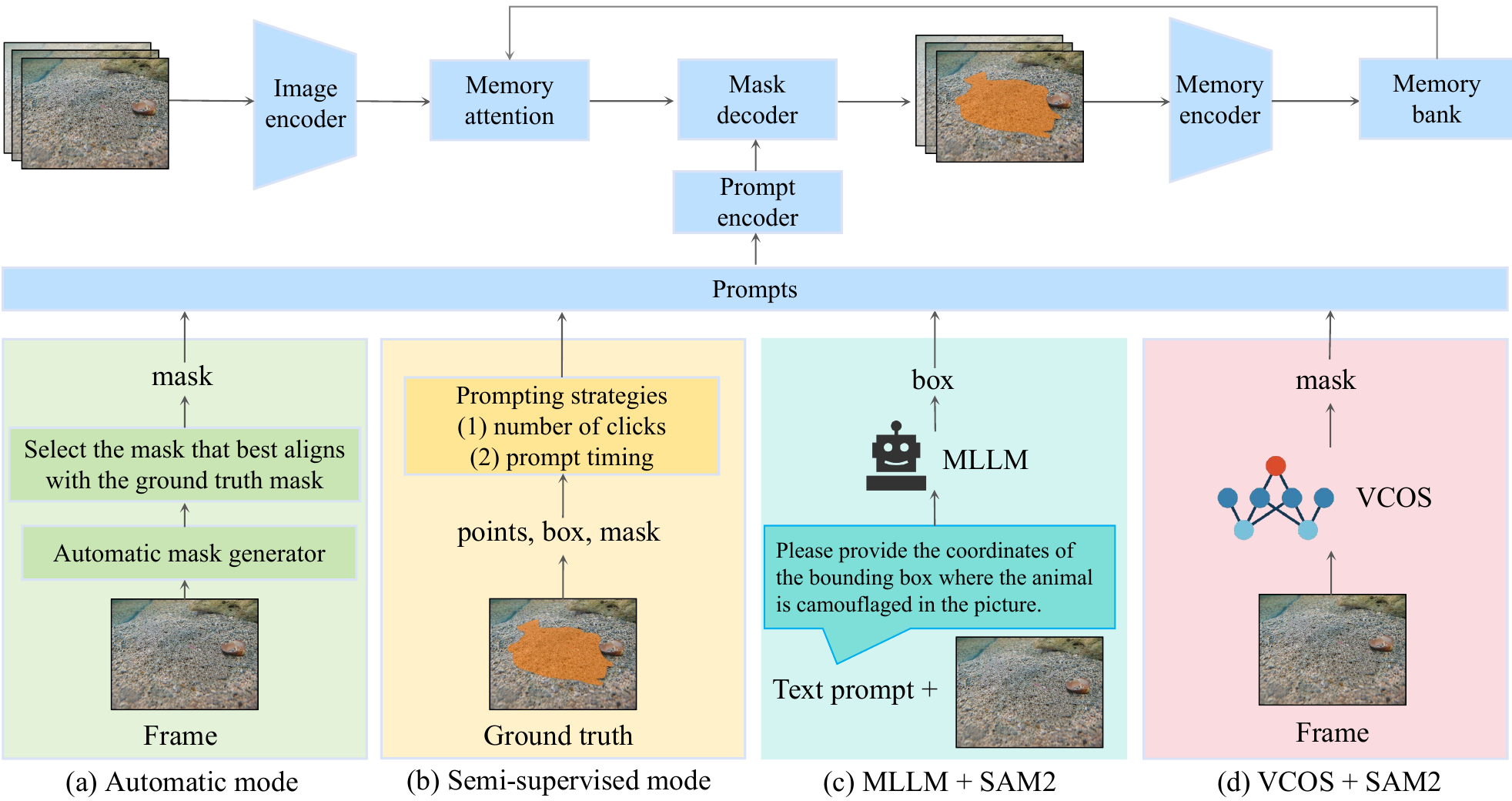}
    \caption{\textbf{Overview of our evaluation framework.} The framework explores different prompts to SAM2 for VCOS. (a) Automatic mode selects the mask (generated by the built-in automatic mask generator) that best aligns with the ground truth mask to serve as the mask prompt. (b) The semi-supervised mode explores different prompt types and different prompt timing based on clicks, boxes, and masks. (c) MLLM + SAM2 utilizes an MLLM to generate bounding box coordinates as the box prompt. (d) VCOS + SAM2 employs a VCOS model to generate a coarse mask as the mask prompt. MLLM: multi-model large language model;  SAM2: segment anything model 2; VCOS: video camouflaged object segmentation.}
    \label{fig:arch}
\end{figure*}

In this section, we present our evaluation framework, as summarized in Fig.~\ref{fig:arch}. We also explain our strategy of fine-tuning SAM2. We first introduce the dataset in Sect. \ref{subsec:datasets} and describe the evaluation metrics in Sect. \ref{subsec:metrics}. In Sect. \ref{subsec:prompting_strategies}, we outline the experimental setup for the automatic and semi-supervised modes of SAM2. Additionally, Sect. \ref{subsec:refine} details the refinement process for results generated by existing MLLM and VCOS methods. Finally, we present the fine-tuning procedure of SAM2 on MoCA-Mask in Sect. \ref{subsec:finetune}.

\subsection{Datasets}
\label{subsec:datasets}
We use two video COD datasets: MoCA-Mask~\cite{cheng2022implicit} and CAD~\cite{bideau2016s}. MoCA-Mask is a densely annotated dataset derived from the moving camouflaged animals (MoCA) dataset~\cite{lamdouar2020betrayed}. It consists of 87 video sequences, with 71 sequences totaling 19,313 frames for training and 16 sequences comprising 3,626 frames for inference, extending the original bounding box annotations to dense segmentation masks on every 5th frames. The camouflaged animal dataset (CAD) includes 9 short video sequences in total that have 181 hand-labeled masks on every 5th frame. Due to some missing ground truth in the CAD test set, we only evaluate the valid video sequences.

\subsection{Metrics}
\label{subsec:metrics}
We use seven common metrics for evaluation including $S$-measure ($S_m$)~\cite{Smeasure}, weighted $F$-measure ($F_\beta^\omega$)~\cite{wFmeasure}, mean absolute error (MAE)~\cite{MAE}, $F$-measure ($F_\beta$)~\cite{Fmeasure}, $E$-measure ($E_m$)~\cite{Emeasure}, mean Dice (mDice) and mean IoU (mIoU).

\subsection{Prompting Strategies}
\label{subsec:prompting_strategies}
We evaluate SAM2's performance in both automatic and semi-supervised modes by utilizing the automatic mask generator and the interactive prompts, respectively. These two modes allow us to thoroughly assess SAM2's flexibility and effectiveness.

\subsubsection{Automatic Mode}
We evaluate SAM2 in automatic mode using its built-in automatic mask generator. In this setup, following the work~\cite{zhang2024evaluation}, SAM2 automatically generates a number of segmentation masks on the first video frame and we select the mask with the highest IoU value compared to the corresponding ground truth. Let $M = \{m_1, m_2, \dots, m_n\}$ represent the set of segmentation masks generated by SAM2 on the first video frame. $G$ represents the ground truth, and $\text{IoU}(m_i, G)$ calculates the IoU between the generated mask $ m_i $ and the ground truth $ G $. The mask $ m^* $ with the highest IoU is selected using Eq.(\ref{eq:1}): 
\begin{equation}
    \label{eq:1}
    m^* = \arg \max_{m_i \in M} \, \text{IoU}(m_i, G)
\end{equation}
The selected mask $ m^* $ is used as the mask prompt, without any manual modification. This ensures that SAM2 operates in an unsupervised manner, relying on its automatic mask generation capabilities to track and segment camouflaged objects throughout the video.

This evaluation aims to assess SAM2's effectiveness in segmenting the camouflaged animals in the video without any user guidance and explore SAM2's potential in such scenarios.

\subsubsection{Semi-supervised Mode}

In the semi-supervised mode, we employ three distinct prompt strategies: click-based, box-based, and mask-based prompts. Each strategy is evaluated across different frames to investigate how prompt types and timing affect SAM2's segmentation performance. For click-based prompts, 1, 3, and 5 foreground clicks (camouflaged animals) are randomly selected from the corresponding ground-truth mask. For box-based and mask-based prompts, we directly use the object's bounding box or mask from the dataset as the prompt.

\subsubsection{Prompt Timing and Frame Selection}

We extend the analysis beyond the first frame by applying prompts at different times in the video sequence. In this experiment, the $0^{th}$, $5^{th}$, $10^{th}$, $-11^{th}$, $-6^{th}$, $-1^{st}$, and middle frame are selected as the prompted frames. Here, the frame index follows indexing rules in the Python list, i.e., $-1^{st}$ represents the last frame. These frames are purposely chosen, from the beginning to the end, allowing us to analyze how prompt timing affects SAM2's performance. Early frames like the $0^{th}$, $5^{th}$, and $10^{th}$ provide insight into how well the model tracks and segments the object from the start, while frames closer to the end, $-11^{th}$, $-6^{th}$, and $-1^{st}$ help evaluate SAM2's ability to handle backward propagation over time. The middle frame is helpful to assess how well SAM2 performs when provided with information at a pivotal moment in the video sequence. This comprehensive selection of frames allows us to analyze how the prompt timing impacts SAM2's segmentation accuracy and robustness when processing videos.

\subsection{Refine MLLMs and VCOS Methods with SAM2}
\label{subsec:refine}
In this experiment, we explore using SAM2 to refine the results generated by the existing MLLMs and VCOS methods.

\subsubsection{Refine MLLMs with SAM2}
The use of MLLMs (Multimodal Large Language Models) with SAM2 is motivated by the limitation that most current MLLMs can only output bounding boxes, but not the segmentation masks. If the approach of using MLLMs-generated bounding boxes as prompts for SAM2 proves effective, it allows for automated identification and segmentation of objects, removing the need for manual prompts. 
% We explore a zero-shot object detection approach using the MLLM to generate bounding box prompts for video segmentation. The process is designed to handle the challenging task of segmenting camouflaged objects in videos.

To be more specific, we employ two large Multimodal LLM models, LLaVA-1.5-7b~\cite{liu2024visual} and Shikra-7b-delta-v1~\cite{chen2023shikra}, in a zero-shot setting to detect camouflaged objects in the first frame of the video. We experimented with several prompts to ask MLLMs to generate the bounding box inspired by~\cite{tang2024chain}, and we finally selected the prompt ``Please provide the coordinates of the bounding box where the animal is camouflaged in the picture''. The input to the model consists of this question prompt alongside the visual input of the first video frame. The model processes this input and outputs the coordinates of a bounding box that is presumed to encapsulate the camouflaged object. The bounding box coordinates generated by the MLLMs serve as the box prompt for SAM2. SAM2 uses these coordinates to guide the segmentation process across the whole video. By leveraging the box prompt, SAM2 is expected to effectively segment the camouflaged object, even in the presence of complex backgrounds and low contrast. 

\subsubsection{Refine VCOS with SAM2}

In this experiment, we focus on refining the camouflaged object masks generated by existing VCOS models using SAM2. Since VCOS models already output the object masks, we explore how SAM2 can enhance the quality of these masks through its advanced segmentation capabilities.

Specifically, we use the object masks produced by the VCOS models as prompts for SAM2. These masks serve as initial rough segmentation, which SAM2 uses to further refine the details. By leveraging SAM2's powerful mask tracking capabilities, we aim to improve the results generated by existing VCOS methods. We use the predicted mask of the middle frame as the prompt.

\subsection{Fine-tune SAM2 on MoCA-Mask}
\label{subsec:finetune}
SAM2 was trained on the large-scale SA-V~\cite{ravi2024sam} dataset which generally contains videos from common scenes. We propose a fine-tuning strategy to adjust SAM2's parameter on the popular MoCA-Mask~\cite{cheng2022implicit} dataset to improve its ability on challenging camouflaged scenes. Specifically, we follow the work of MedSAM2~\cite{ma2024segmentmedicalimagesvideos}, fine-tune the image encoder, the mask decoder, and both, with freezing other parameters to maintain generalization since SAM2 was initially trained on a much larger dataset. We followed the approach of MedSAM2 and added a shift of the bounding box used during training. This selective fine-tuning focuses on adapting SAM2 to the specific challenges of camouflaged object detection, where precise feature encoding and decoding are critical. We employ the AdamW~\cite{loshchilov2017decoupled} optimizer with the combination of Dice loss and Binary Cross-Entropy (BCE) loss to achieve accurate segmentation of camouflaged objects. We use the learning rate (\(lr=1e-8\)) and weight decay (\(wd=0.01\)). The overall loss function is computed as a sum of the segmentation loss and the BCE loss: 
\begin{equation}
    \mathcal{L} = \mathcal{L}_{seg} + \mathcal{L}_{ce}
\end{equation}

\section{Results}
In this section, we present the results of our evaluation and adaptation of SAM2. We compare SAM2's performance with state-of-the-art methods and analyze the effects of different prompting strategies, including variations in the number of clicks and prompt timing. We also examine the effectiveness of integrating MLLMs and existing VCOS methods with SAM2 and fine-tuning SAM2 on the MoCA-Mask dataset. We conduct all experiments on a single NVIDIA RTX A6000 with the \texttt{torch.float32} precision.

\begin{table*}[t]
\caption{State-of-the-art comparisons on the CAD dataset~\cite{bideau2016s}. FPS: inference speed; Param.: parameters; MAE: mean absolute error. The best results are shown in bold.}
\label{tab:res_comparison_cad}
\centering
\resizebox{\textwidth}{!}{%
\begin{tabular}{c|c|c|c|ccccccc}
\toprule
\multirow{2}{*}{Model} & \multirow{2}{*}{Backbone} & \multirow{2}{*}{FPS} & \multirow{2}{*}{Param.} & \multicolumn{7}{c}{CAD} \\ \cmidrule{5-11} 
 &  &  &  & $S_m \uparrow$ & $F_\beta^\omega \uparrow$ & MAE $\downarrow$ & $F_\beta \uparrow$ & $E_m \uparrow$ & mDice $\uparrow$ & mIoU $\uparrow$ \\ \midrule
EGNet~\cite{zhao2019egnet} & ResNet-50 & 5.7 & 111.7M & 0.619 & 0.298 & 0.044 & 0.350 & 0.666 & 0.324 & 0.243 \\
BASNet~\cite{qin2019basnet} & ResNet-50 & 37.4 & 87.1M & 0.639 & 0.349 & 0.054 & 0.394 & 0.773 & 0.393 & 0.293 \\
CPD~\cite{wu2019cascaded} & ResNet-50 & 43.7 & 47.9M & 0.622 & 0.289 & 0.049 & 0.357 & 0.667 & 0.330 & 0.239 \\
PraNet~\cite{fan2020pranet} & ResNet-50 & 41.9 & 32.6M & 0.629 & 0.352 & 0.042 & 0.397 & 0.763 & 0.378 & 0.290 \\
SINet~\cite{fan2020Camouflage} & ResNet-50 & 56.5 & 48.9M & 0.636 & 0.346 & 0.041 & 0.395 & 0.775 & 0.381 & 0.283 \\
SINet-V2~\cite{fan2021concealed} & Res2Net-50 & 38.3 & 27.0M & 0.653 & 0.382 & 0.039 & 0.432 & 0.762 & 0.413 & 0.318 \\
PNS-Net~\cite{ji2021progressively} & ResNet-50 & 31.7 & 142.9M & 0.655 & 0.325 & 0.048 & 0.417 & 0.673 & 0.384 & 0.290 \\
RCRNet~\cite{yan2019semi} & ResNet-50 & 58.1 & 53.8M & 0.627 & 0.287 & 0.048 & 0.328 & 0.666 & 0.309 & 0.229 \\
MG~\cite{yang2021self} & VGG & 272.9 & 4.8M & 0.594 & 0.336 & 0.059 & 0.375 & 0.692 & 0.368 & 0.268 \\
SLT-Net-LT~\cite{cheng2022implicit} & PVTv2-B5 & 35.6 & 82.3M & 0.696 & 0.481 & 0.030 & 0.524 & 0.845 & 0.493 & 0.402 \\
ZoomNeXt~\cite{pang2024zoomnext} & PVTv2-B5 & 26.5 & 84.8M & 0.757 & 0.593 & 0.020 & 0.631 & 0.865 & 0.599 & 0.510 \\ \midrule

SAM2~\cite{ravi2024sam} (1-click) & Hiera-T & 47.2 & 38.9M & 0.754 & 0.612 & 0.033 & 0.652 & 0.813 & 0.622 & 0.515 \\

SAM2~\cite{ravi2024sam} (box) & Hiera-T & 47.2 & 38.9M & 0.852 & 0.803 & 0.016 & 0.826 & 0.951 & 0.810 & 0.694 \\

SAM2~\cite{ravi2024sam} (mask) & Hiera-T & 47.2 & 38.9M & 0.887 & 0.845 & 0.010 & 0.856 & 0.971 & 0.852 & 0.756 \\ 
\midrule

SAM2~\cite{ravi2024sam} (1-click) & Hiera-S & 43.3 & 46.0M & 0.742 & 0.587 & 0.031 & 0.641 & 0.755 & 0.586 & 0.486 \\

SAM2~\cite{ravi2024sam} (box) & Hiera-S & 43.3 & 46.0M & 0.879 & 0.840 & 0.011 & 0.854 & 0.969 & 0.844 & 0.743 \\

SAM2~\cite{ravi2024sam} (mask) & Hiera-S & 43.3 & 46.0M & 0.888 & 0.854 & 0.010 & 0.865 & 0.975 & 0.859 & 0.766 \\
\midrule

SAM2~\cite{ravi2024sam} (1-click) & Hiera-B+ & 34.8 & 80.8M & 0.775 & 0.669 & 0.027 & 0.736 & 0.823 & 0.671 & 0.546 \\

SAM2~\cite{ravi2024sam} (box) & Hiera-B+ & 34.8 & 80.8M & 0.866 & 0.816 & 0.012 & 0.835 & 0.960 & 0.821 & 0.714 \\

SAM2~\cite{ravi2024sam} (mask) & Hiera-B+ & 34.8 & 80.8M & 0.883 & 0.842 & 0.009 & 0.856 & 0.966 & 0.847 & 0.752 \\
\midrule

SAM2~\cite{ravi2024sam} (1-click) & Hiera-L & 24.2 & 224.4M & 0.749 & 0.592 & 0.028 & 0.647 & 0.738 & 0.592 & 0.497 \\

SAM2~\cite{ravi2024sam} (box) & Hiera-L & 24.2 & 224.4M & 0.872 & 0.836 & 0.012 & 0.853 & 0.969 & 0.839 & 0.735 \\

SAM2~\cite{ravi2024sam} (mask) & Hiera-L & 24.2 & 224.4M & \textbf{0.893} & \textbf{0.863} & \textbf{0.008} & \textbf{0.873} & \textbf{0.976} & \textbf{0.864} & \textbf{0.775} \\\bottomrule
\end{tabular}%
}

\end{table*}

\begin{table*}[t]
\centering
\caption{State-of-the-art comparisons on the MoCA-Mask dataset~\cite{cheng2022implicit}.}
\label{tab:res_comparison_moca}
\resizebox{\textwidth}{!}{%
\begin{tabular}{c|c|c|c|ccccccc}
\toprule
\multirow{2}{*}{Model} & \multirow{2}{*}{Backbone} & \multirow{2}{*}{FPS} & \multirow{2}{*}{Param.} & \multicolumn{7}{c}{MoCA-Mask} \\ \cmidrule{5-11} 
 &  &  &  & $S_m \uparrow$ & $F_\beta^\omega \uparrow$ & MAE $\downarrow$ & $F_\beta \uparrow$ & $E_m \uparrow$ & mDice $\uparrow$ & mIoU $\uparrow$ \\ \midrule
EGNet~\cite{zhao2019egnet} & ResNet-50 & 5.7 & 111.7M & 0.547 & 0.110 & 0.035 & 0.136 & 0.574 & 0.143 & 0.096 \\
BASNet~\cite{qin2019basnet} & ResNet-50 & 37.4 & 87.1M & 0.561 & 0.154 & 0.042 & 0.173 & 0.598 & 0.190 & 0.137 \\
CPD~\cite{wu2019cascaded} & ResNet-50 & 43.7 & 47.9M & 0.561 & 0.121 & 0.041 & 0.152 & 0.613 & 0.162 & 0.113 \\
PraNet~\cite{fan2020pranet} & ResNet-50 & 41.9 & 32.6M & 0.614 & 0.266 & 0.030 & 0.296 & 0.674 & 0.311 & 0.234 \\
SINet~\cite{fan2020Camouflage} & ResNet-50 & 56.5 & 48.9M & 0.598 & 0.231 & 0.028 & 0.256 & 0.699 & 0.277 & 0.202 \\
SINet-V2~\cite{fan2021concealed} & Res2Net-50 & 38.3 & 27.0M & 0.588 & 0.204 & 0.031 & 0.229 & 0.642 & 0.245 & 0.180 \\
PNS-Net~\cite{ji2021progressively} & ResNet-50 & 31.7 & 142.9M & 0.526 & 0.059 & 0.035 & 0.084 & 0.530 & 0.084 & 0.054 \\
RCRNet~\cite{yan2019semi} & ResNet-50 & 58.1 & 53.8M & 0.555 & 0.138 & 0.033 & 0.159 & 0.527 & 0.171 & 0.116 \\
MG~\cite{yang2021self} & VGG & 272.9 & 4.8M & 0.530 & 0.168 & 0.067 & 0.195 & 0.561 & 0.181 & 0.127 \\
SLT-Net-LT~\cite{cheng2022implicit} & PVTv2-B5 & 35.6 & 82.3M & 0.631 & 0.311 & 0.027 & 0.331 & 0.759 & 0.360 & 0.272 \\
ZoomNeXt~\cite{pang2024zoomnext} & PVTv2-B5 & 26.5 & 84.8M & 0.734 & 0.476 & 0.010 & 0.497 & 0.736 & 0.497 & 0.422 \\ \midrule

SAM2~\cite{ravi2024sam} (1-click) & Hiera-T & 47.2 & 38.9M & 0.680 & 0.523 & 0.074 & 0.556 & 0.775 & 0.532 & 0.444 \\
SAM2~\cite{ravi2024sam} (box) & Hiera-T & 47.2 & 38.9M & 0.813 & 0.697 & 0.006 & 0.707 & 0.895 & 0.726 & 0.621 \\
SAM2~\cite{ravi2024sam} (mask) & Hiera-T & 47.2 & 38.9M & 0.846 & 0.760 & \textbf{0.004} & 0.770 & 0.918 & 0.771 & 0.679 \\ \midrule

SAM2~\cite{ravi2024sam} (1-click) & Hiera-S & 43.3 & 46.0M & 0.719 & 0.534 & 0.009 & 0.567 & 0.757 & 0.546 & 0.456 \\
SAM2~\cite{ravi2024sam} (box) & Hiera-S & 43.3 & 46.0M & 0.826 & 0.715 & 0.006 & 0.724 & 0.887 & 0.738 & 0.637 \\
SAM2~\cite{ravi2024sam} (mask) & Hiera-S & 43.3 & 46.0M & \textbf{0.862} & 0.791 & \textbf{0.004} & 0.801 & \textbf{0.946} & 0.802 & 0.706 \\ \midrule

SAM2~\cite{ravi2024sam} (1-click) & Hiera-B+ & 34.8 & 80.8M & 0.752 & 0.622 & 0.008 & 0.666 & 0.804 & 0.639 & 0.523 \\
SAM2~\cite{ravi2024sam} (box) & Hiera-B+ & 34.8 & 80.8M & 0.819 & 0.707 & 0.007 & 0.717 & 0.909 & 0.730 & 0.631 \\
SAM2~\cite{ravi2024sam} (mask) & Hiera-B+ & 34.8 & 80.8M & 0.846 & 0.768 & \textbf{0.004} & 0.780 & 0.915 & 0.776 & 0.684 \\
\midrule

SAM2~\cite{ravi2024sam} (1-click) & Hiera-L & 24.2 & 224.4M & 0.787 & 0.658 & 0.006 & 0.679 & 0.844 & 0.670 & 0.575 \\
SAM2~\cite{ravi2024sam} (box) & Hiera-L & 24.2 & 224.4M & 0.832 & 0.729 & 0.006 & 0.733 & 0.926 & 0.748 & 0.653 \\
SAM2~\cite{ravi2024sam} (mask) & Hiera-L & 24.2 & 224.4M & \textbf{0.862} & \textbf{0.796} & \textbf{0.004} & \textbf{0.806} & 0.943 & \textbf{0.806} & \textbf{0.709} \\ \bottomrule
\end{tabular}%
}

\end{table*}

\subsection{Comparison with State-of-the-art VCOS Methods}

We compare the results obtained by prompting SAM2 using 1-click, box, and mask at the first video frame, with existing state-of-the-art methods, in Tab.~\ref{tab:res_comparison_cad} and Tab.~\ref{tab:res_comparison_moca}. It is observed that SAM2 with semi-supervised mode outperforms current SOTA models in the VCOS task, where prompts guide the segmentation process. This naturally gives SAM2 an advantage, as the use of interactive prompts allows for better adaptability compared to fully automated models. These results highlight the effectiveness of SAM2 for camouflaged video segmentation.

Notably, SAM2 demonstrates a strong balance between model size, inference speed (FPS), and segmentation performance, making it highly adaptable for VCOS. Compared to models such as EGNet (111.7M), BASNet (87.1M), and PNS-Net (142.9M), SAM2 achieves superior performance with significantly fewer parameters, particularly in its Hiera-T (38.9M) and Hiera-S (46.0M) versions. Inference speed is another critical factor, with Hiera-T achieving 47.2 FPS and Hiera-S running at 43.3 FPS, outperforming many state-of-the-art models, including SINet-V2 (38.3 FPS) and SLT-Net-LT (35.6 FPS). While SAM2 of larger size versions like Hiera-B+ (80.8M) further enhances segmentation accuracy, they come at the cost of increased computational complexity and slower FPS. Nevertheless, SAM2 achieves higher segmentation accuracy than transformer-based models such as ZoomNeXt (26.5 FPS) despite being faster and more lightweight, demonstrating its efficiency. These results highlight that Hiera-T and Hiera-S offer the best trade-off between speed and model size, making SAM2 a strong candidate for real-time and resource-constrained VCOS scenarios.

\subsection{Effect of Prompting Strategies}
We compare results obtained from various prompting strategies as mentioned in Sect. \ref{subsec:prompting_strategies}. From the results, we have three observations: 1) Mask-based prompt results in the best segmentation result, compared with click and box prompts; 2) Increasing the number of clicks significantly improves the segmentation result; 3) Prompting at the middle frame of the video generally gives better performance than prompting at other frames and results vary a lot when prompting at different times.

\subsubsection{Comparisons among Different Prompting Strategies}
Tab.~\ref{tab:res_comparison_cad} and Tab.~\ref{tab:res_comparison_moca} presents a comprehensive comparison of SAM2's performance using different prompting strategies (1-click, box, and mask) across various model sizes on the CAD and MoCA-Mask datasets. The results clearly demonstrate the superiority of mask-based prompts over both 1-click and box-based strategies. This result is consistent with our intuition: the more detailed the prompt, the better the segmentation. For instance, on the CAD dataset, SAM2 with mask prompting achieves the highest mIoU of 0.775 for the large model (Hiera-L), outperforming both 1-click (mIoU 0.497) and box-based prompting (mIoU 0.735). Similar trends are observed on MoCA-Mask with Hiera-L, where mask prompting achieves 0.709 in mIoU, compared to 0.575 for 1-click prompting and 0.653 for box-based prompting.

\subsubsection{Impact of the Number of Clicks for Point Prompting}
We examine the impact of different numbers of clicks (1, 3, and 5) used as prompts at the middle frame on segmentation performance, across various model sizes. The results on the MoCA-Mask dataset are shown in Tab.~\ref{tab:click_performance}. For the SAM2-L model, increasing the number of clicks leads to a consistent performance improvement; the 5-click prompt achieves the highest scores (i.e., $S_m$=0.831, mDice=0.748, and mIoU=0.652) compared to both the 1-click and 3-click settings.
In contrast, for the other three models (SAM2-B+, SAM2-S, and SAM2-T), the 3-click outperforms the 5-click settings, although the performance differences are minimal. For instance, the SAM2-B+ model achieves a mIoU of 0.624 with 3-click prompts against 0.606 with 5-click prompts. 
Although the trends between the 3-click and 5-click settings vary across different model sizes, both settings consistently outperform the 1-click settings. This indicates that incorporating additional point prompts generally provides more informative guidance, thereby enhancing the segmentation performance. In conclusion, employing multiple clicks leads to improved results compared to using a single click, highlighting the benefit of human guidance in the VCOS task.

\begin{table*}[t]
\centering
\caption{Click-based prompt performance on the middle frame with different click counts on the MoCA-Mask dataset~\cite{cheng2022implicit}.}
\label{tab:click_performance}
\resizebox{\textwidth}{!}{%
\begin{tabular}{c|c|ccccccc}
\toprule
~~~\multirow{2}{*}{\textbf{Model}}~~~ & ~~~\multirow{2}{*}{\textbf{Click Count}}~~~ & \multicolumn{7}{c}{\textbf{MoCA-Mask}} \tabularnewline \cmidrule{3-9} 
 &  & ~~~$S_m \uparrow$~~~ & ~~~$F_\beta^\omega \uparrow$~~~ & ~~~MAE $\downarrow$~~~ & ~~~$F_\beta \uparrow$~~~ & ~~~$E_m \uparrow$~~~ & ~~~mDice $\uparrow$~~~ & ~~~mIoU $\uparrow$~~~ \\ \midrule

\multirow{3}{*}{SAM2-L}
 & 1 & 0.746 & 0.619 & 0.064 & 0.635 & 0.810 & 0.634 & 0.548 \\
 & 3 & 0.785 & 0.702 & 0.063 & 0.713 & 0.873 & 0.718 & 0.624 \\
 & 5 & \textbf{0.831} & \textbf{0.729} & \textbf{0.008} & \textbf{0.736} & \textbf{0.909} & \textbf{0.748} & \textbf{0.652} \\ \midrule

 \multirow{3}{*}{SAM2-B+} 
 & 1 & 0.738 & 0.604 & 0.065 & 0.617 & 0.821 & 0.617 & 0.533 \\
 & 3 & \textbf{0.812} & \textbf{0.701} & \textbf{0.007} & \textbf{0.712} & \textbf{0.886} & \textbf{0.722} & \textbf{0.624} \\
 & 5 & 0.808 & 0.677 & 0.009 & 0.686 & 0.870 & 0.703 & 0.606 \\\midrule

 \multirow{3}{*}{SAM2-S} 
 & 1 & 0.748 & 0.611 & 0.058 & 0.626 & 0.808 & 0.621 & 0.538 \\
 & 3 & 0.783 & \textbf{0.682} & 0.053 & \textbf{0.693} & \textbf{0.873} & \textbf{0.703} & \textbf{0.606} \\
 & 5 & \textbf{0.788} & 0.671 & \textbf{0.047} & 0.680 & 0.866 & 0.699 & 0.599 \\ 
\midrule

 \multirow{3}{*}{SAM2-T} 
 & 1 & 0.768 & 0.614 & \textbf{0.007} & 0.638 & 0.820 & 0.629 & 0.536 \\
 & 3 & 0.787 & \textbf{0.649} & 0.011 & \textbf{0.661} & 0.872 & \textbf{0.676} & \textbf{0.576} \\
 & 5 & \textbf{0.793} & 0.644 & 0.012 & 0.654 & \textbf{0.881} & 0.675 & 0.572 \\ 
 
 \bottomrule
\end{tabular}%
}

\end{table*}

\subsubsection{Impact of Prompt Timing}
We evaluate the effect of prompt timing on the performance using click-based, box-based, and mask-based prompts on the small version of the SAM2 using the MoCA-Mask dataset. Tab.~\ref{tab:timing_performance} presents the results for different prompt timings across all three prompt strategies. In general, the results show that applying the prompt on the middle frame yields the best segmentation performance across all strategies. For instance, using the click-based prompts in the middle frame gives a mIoU of 0.538, which is higher than prompting at other times. Similarly, for the box-based and mask-based prompts, the middle frame provides the best results in most evaluation metrics. The trend is particularly obvious for the mask-based prompts where the middle frame achieves a mIoU of 0.719, the highest among all experiments. These results suggest that prompt timing is a critical factor in achieving optimal segmentation performance.

\begin{table*}[t]
\centering
\caption{Performance of the SAM2-S with different prompt timings on the MoCA-Mask dataset~\cite{cheng2022implicit}.}
\label{tab:timing_performance} 
\resizebox{\textwidth}{!}{%
\begin{tabular}{c|c|ccccccc}
\toprule
~~~\multirow{2}{*}{\textbf{Prompt Type}}~~~ & ~~~\multirow{2}{*}{\textbf{Frame}}~~~ & \multicolumn{7}{c}{\textbf{MoCA-Mask}} \\ \cmidrule{3-9} 
 &  & ~~~$S_m \uparrow$~~~ & ~~~$F_\beta^\omega \uparrow$~~~ & ~~~MAE $\downarrow$~~~ & ~~~$F_\beta \uparrow$~~~ & ~~~$E_m \uparrow$~~~ & ~~~mDice $\uparrow$~~~ & ~~~mIoU $\uparrow$~~~ \\ \midrule
\multirow{7}{*}{1-click} 
 & 0 & 0.719 & 0.534 & 0.009 & 0.567 & 0.757 & 0.546 & 0.456 \\
 & 5 & \textbf{0.766} & 0.597 & \textbf{0.007} & 0.605 & 0.795 & 0.611 & 0.532  \\
 & 10 & 0.732 & 0.548 & 0.009 & 0.568 & 0.763 & 0.561 & 0.479 \\
 & -11 & 0.732 & 0.566 & 0.054 & 0.585 & 0.758 & 0.578 & 0.497 \\
 & -6 & 0.746 & 0.603 & 0.054 & \textbf{0.632} & 0.779 & 0.618 & 0.524 \\
 & -1 & 0.742 & 0.601 & 0.067 & 0.623 & 0.773 & 0.616 & 0.531  \\
 & middle & 0.748 & \textbf{0.611} & 0.058 & 0.626 & \textbf{0.808} & \textbf{0.621} & \textbf{0.538} \\ \midrule
 
\multirow{7}{*}{Box} 
 & 0 & 0.826 & 0.715 & 0.006 & 0.724 & 0.887 & 0.738 & 0.637 \\
 & 5 & 0.816 & 0.698 & 0.006 & 0.705 & 0.871 & 0.718 & 0.623  \\
 & 10 & 0.810 & 0.699 & \textbf{0.005} & 0.715 & 0.878 & 0.717 & 0.614 \\
 & -11 & 0.832 & 0.719 & \textbf{0.005} & 0.723 & 0.899 & 0.737 & 0.643 \\
 & -6 & \textbf{0.834} & 0.731 & \textbf{0.005} & 0.735 & 0.898 & 0.747 & 0.653 \\
 & -1 & 0.831 & 0.718 & \textbf{0.005} & 0.722 & 0.891 & 0.739 & 0.644  \\
 & middle & \textbf{0.834} & \textbf{0.738} & \textbf{0.005} & \textbf{0.744} & \textbf{0.906} & \textbf{0.753} & \textbf{0.657} \\
 \midrule

\multirow{7}{*}{Mask} 
 & 0 & 0.862 & 0.791 & \textbf{0.004} & 0.801 & 0.946 & 0.802 & 0.706 \\
 & 5 & 0.850 & 0.770 & \textbf{0.004} & 0.780 & 0.930 & 0.781 & 0.688  \\
 & 10 & 0.854 & 0.777 & \textbf{0.004} & 0.787 & 0.932 & 0.787 & 0.693 \\
 & -11 & 0.863 & 0.784 & \textbf{0.004} & 0.790 & 0.936 & 0.798 & 0.704 \\
 & -6 & 0.862 & 0.787 & \textbf{0.004} & 0.797 & 0.939 & 0.799 & 0.703 \\
 & -1 & 0.860 & 0.781 & \textbf{0.004} & 0.790 & 0.934 & 0.797 & 0.699  \\
 & middle & \textbf{0.873} & \textbf{0.803} & \textbf{0.004} & \textbf{0.810} & \textbf{0.954} & \textbf{0.813} & \textbf{0.719} \\
 \bottomrule
\end{tabular}% 
} 

\end{table*}

\subsection{Automatic Mode Results}

\begin{figure*}[t]
    \centering
    \includegraphics[width=\linewidth]{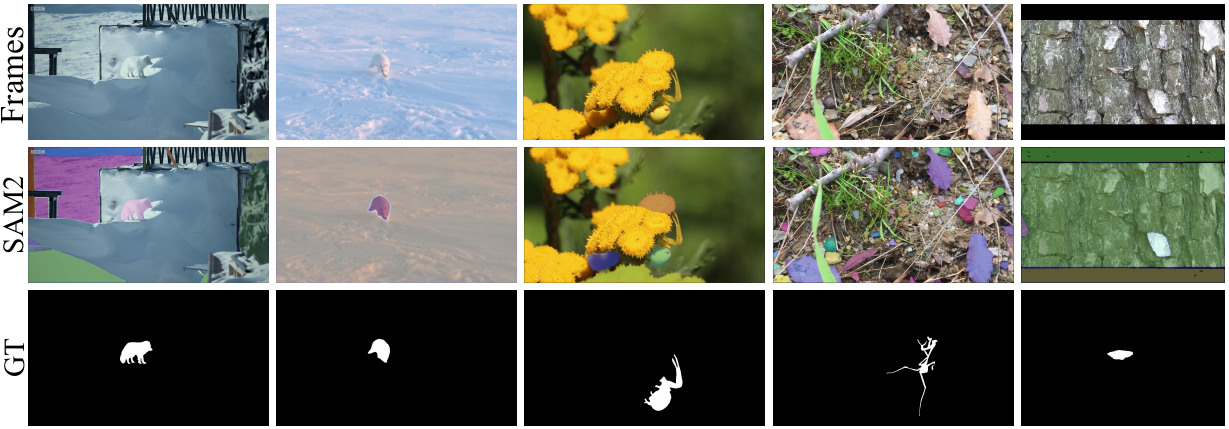}
    \caption{\textbf{Visualization of masks generated by Automatic mode of SAM2 on MoCA-Mask.} From \textit{top} to \textit{bottom}: the input frames, masks generated in automatic mode, and the ground truths. SAM2 can generate multiple masks (shown in different colors) for this mode. Best viewed in color.}
    \label{fig:auto_mode_vis}
\end{figure*}

\begin{table*}[t]
\centering
\caption{Automatic mode performance of SAM2 on the MoCA-Mask dataset~\cite{cheng2022implicit}.}
\label{tab:auto}
\resizebox{\textwidth}{!}{%
\begin{tabular}{c|ccccccc}
\toprule
~~~~\textbf{Size}~~~~ & ~~~~~$S_m \uparrow$ ~~~~~ & ~~~~~ $F_\beta^\omega \uparrow$~~~~~ & ~~~~~MAE $\downarrow$~~~~~ & ~~~~~$F_\beta \uparrow$~~~~~ & ~~~~~$E_m \uparrow$~~~~~ & ~~~~~mDice $\uparrow$~~~~~ & ~~~~~mIoU $\uparrow$~~~~~ \\ \midrule
SAM2-L & 0.466 & 0.157 & 0.190 & 0.165 & 0.584 & 0.157 & 0.138 \\
SAM2-B+ & 0.468 & 0.137 & \textbf{0.141} & 0.150 & 0.561 & 0.138 & 0.113 \\
SAM2-S & \textbf{0.497} & \textbf{0.201} & 0.148 & \textbf{0.214} & 0.608 & \textbf{0.202} & \textbf{0.174} \\
SAM2-T & 0.495 & 0.166 & 0.154 & 0.167 & \textbf{0.634} & 0.167 & 0.151 \\ \bottomrule

\end{tabular}
}

\end{table*}

The results of using SAM2 in automatic mode on the MoCA-Mask dataset are shown in Tab.~\ref{tab:auto}. It can be observed that the performance is notably lower compared to semi-supervised mode with guided prompts. Fig.~\ref{fig:auto_mode_vis} presents some visualizations of automatically generated masks for the first frame in each video, using Hiera-L as the backbone. From left to right, SAM2 successfully segments the camouflaged objects in the first two examples. However, in the middle example, it only partially segments the camouflaged object, and in the last two examples, SAM2 fails to segment the camouflaged object entirely. SAM2 is optimized for natural scenes rather than camouflaged environments, so it is more difficult for SAM2 to determine camouflaged objects without any guidance (prompts), especially when the camouflaged objects are not clearly distinguishable from the background.

\subsection{Refine MLLMs and VCOS with SAM2}
\subsubsection{Refine MLLMs with SAM2}
\begin{table*}[t]
\centering
\caption{Performance comparison of LLaVA+SAM2 and Shikra+SAM2 models on VCOS on the MoCA-Mask dataset~\cite{cheng2022implicit}.}
\label{tab:MLLM}
\resizebox{\textwidth}{!}{%
\begin{tabular}{c|ccccccc}
\toprule
~~~~~~\textbf{Model}~~~~~ & ~~~~$S_m \uparrow$ ~~~~ & ~~~~ $F_\beta^\omega \uparrow$~~~~ & ~~~~MAE $\downarrow$~~~~ & ~~~~$F_\beta \uparrow$~~~~ & ~~~~$E_m \uparrow$~~~~ & ~~~~mDice $\uparrow$~~~~ & ~~~~mIoU $\uparrow$~~~~ \\ \midrule
LLaVA+SAM2-L & \textbf{0.624} & \textbf{0.315} & \textbf{0.046} & \textbf{0.317} & \textbf{0.688} & \textbf{0.334} & \textbf{0.291} \\ 
LLaVA+SAM2-B+ & 0.552 & 0.213 & 0.080 & 0.227 & 0.657 & 0.249 & 0.190 \\ 
LLaVA+SAM2-S & 0.518 & 0.159 & 0.090 & 0.173 & 0.558 & 0.179 & 0.137 \\ 
LLaVA+SAM2-T & 0.605 & 0.289 & 0.070 & 0.298 & 0.687 & 0.329 & 0.266 \\ 
\midrule

Shikra+SAM2-L & \textbf{0.502} & 0.146 & \textbf{0.107} & 0.155 & \textbf{0.590} & 0.157 & 0.124 \\ 
Shikra+SAM2-B+ & 0.490 & \textbf{0.193} & 0.170 & \textbf{0.204} & 0.537 & \textbf{0.208} & \textbf{0.171} \\ 
Shikra+SAM2-S & 0.444 & 0.107 & 0.190 & 0.115 & 0.530 & 0.114 & 0.090 \\ 
Shikra+SAM2-T & 0.473 & 0.119 & 0.139 & 0.130 & 0.532 & 0.133 & 0.101 \\  

\bottomrule
\end{tabular}
}

\end{table*}

In our experiment, we utilize two large Multimodal LLM models, LLaVA-1.5-7b~\cite{liu2024visual} and Shikra-7b-delta-v1~\cite{chen2023shikra}, in combination with SAM2 for video segmentation. The results are presented in Tab.~\ref{tab:MLLM}. It shows that the performance is unsatisfactory. For instance, the mIoU and mDice scores for both LLaVA+SAM2 and Shikra+SAM2 models are relatively low, with the 
large model sizes achieving mIoU values of 0.291 and 0.124, respectively. The poor performance can be attributed to the heavy reliance of SAM2 on accurate bounding box detection by the MLLMs in the first frame. When the MLLMs fail to generate the correct bounding box, it severely impacts segmentation in the subsequent frames, leading to poor segmentation masks. This highlights the crucial role of accurate object detection in the initial frame. Although the current MLLM+SAM2 framework does not achieve strong performance on the VCOS problem, it presents a promising heuristic approach for fully automating SAM2-based segmentation without requiring manual prompts. For example, LLaVA+SAM2-L, while performing worse than the semi-supervised mode of SAM2, still surpasses many existing methods, as we listed in Tab.~\ref{tab:res_comparison_moca}. 

The integration of MLLMs inevitably introduces additional computational overhead; however, in our approach, this impact is minimal. Specifically, we utilize MLLMs only for the first frame of the video sequence to generate the initial bounding box prompt rather than processing each frame individually. Once the prompt is obtained, SAM2 operates independently throughout the remaining frames, ensuring efficient segmentation and tracking without further reliance on MLLM inference. As a result, the FPS degradation and latency introduced by MLLM processing are negligible over the course of a long video sequence. Since the MLLM inference occurs only once at the beginning, its computational cost is amortized across the entire sequence, making the overhead practically insignificant for real-world applications.

However, due to the inherent difficulty in detecting camouflaged animals, MLLMs often generate imprecise bounding box coordinates, leading to incorrect prompts for SAM2. These inaccurate prompts, in turn, cause segmentation failures and disrupt object tracking across frames, reducing the effectiveness of the framework.

\subsubsection{Refine VCOS with SAM2}
\begin{table*}[t]
\centering
\caption{Refinement of TSP-SAM with SAM2 on camouflaged object segmentation.}
\label{tab:refine_tsp_sam}
\resizebox{\textwidth}{!}{%
\begin{tabular}{c|ccccccc}
\toprule
~~~~~~\textbf{Model}~~~~~~ & ~~~~~~$S_m \uparrow$ ~~~~~~ & ~~~~~~ $F_\beta^\omega \uparrow$~~~~~~ & ~~~~~~MAE $\downarrow$~~~~~~ & ~~~~~~$E_m \uparrow$~~~~~~ & ~~~~~~mDice $\uparrow$~~~~~~ & ~~~~~~mIoU $\uparrow$~~~~~~ \\ \midrule
TSP-SAM & 0.689 & 0.444 & 0.008 & 0.808 & \textbf{0.458} & 0.388 \\ \midrule
SAM2-L & \textbf{0.696} & \textbf{0.448} & 0.008 & \textbf{0.825} & 0.457 & \textbf{0.408} \\ 
SAM2-B+ & 0.695 & 0.444 & \textbf{0.007} & 0.821 & 0.451 & 0.403 \\ 
SAM2-S & 0.693 & 0.441 & 0.008 & 0.824 & 0.450 & 0.399 \\
SAM2-T & 0.689 & 0.433 & 0.008 & 0.812 & 0.442 & 0.392 \\ 
\bottomrule
\end{tabular}
}

\end{table*}
In this experiment, we focus on refining the segmentation masks generated by the TSP-SAM~\cite{hui2024endow} model for VCOS using SAM2. The initial masks produced by TSP-SAM are used as prompts to SAM2 for refining details. The TSP-SAM model segments each frame based on the preceding frames, progressively improving segmentation with each step. Therefore, we assume that the last frame in the video contains the most refined segmentation result, making it the best candidate for further refinement by SAM2, so we prompt the mask of the last frame of the video sequence for SAM2 refinement.

Tab.~\ref{tab:refine_tsp_sam} presents the results of this refinement process across different model sizes (large, base plus, small, and tiny). Compared to the baseline (TSP-SAM), SAM2 clearly shows improvements on most metrics. For instance, using the large model for the refinement gives an $S_m$ of 0.696 and mIoU of 0.408, marking improvements in structure and segmentation quality. Similar trends are observed across other models, with the tiny model also showing an increase in mIoU improving from 0.388 to 0.392. These results highlight that SAM2's advanced capabilities can clearly enhance the performance of VCOS models.

\subsection{Fine-tune SAM2 on MoCA-Mask}

\begin{table*}[ht]
\centering
\caption{Performance of fine-tuning SAM2-T on the MoCA-Mask dataset~\cite{cheng2022implicit}. Note that \fire\ represents that the weights are updated during fine-tuning.}
\label{tab:res_finetune}
\resizebox{\textwidth}{!}{%
\begin{tabular}{c|ccccccc}
\toprule
~~~~~~\textbf{Method}~~~~~~ & ~~$S_m \uparrow$ ~~ & ~~ $F_\beta^\omega \uparrow$~~ & ~~MAE $\downarrow$~~ & ~~$F_\beta \uparrow$~~ & ~~$E_m \uparrow$~~ & ~~mDice $\uparrow$~~ & ~~mIoU $\uparrow$~~ \\ \midrule
SAM2-T(baseline) & 0.815 & 0.699 & 0.006 & 0.710 & 0.895 & 0.728 & 0.623 \\ \midrule
SAM2-T(Image Encoder{\fire}) & 0.831 & 0.724 & \textbf{0.005} & 0.730 & 0.907 & 0.754 & 0.651 \\ 
SAM2-T(Mask Decoder\fire)  & 0.818 & 0.706 & 0.006 & 0.716 & 0.899 & 0.733 & 0.629 \\ 
SAM2-T(Image Encoder\fire +Mask Decoder\fire) & \textbf{0.832} & \textbf{0.726} & \textbf{0.005} & \textbf{0.733} & \textbf{0.908} & \textbf{0.756} & \textbf{0.652} \\ 
\bottomrule
\end{tabular}
}

\end{table*}

\begin{table*}[ht]
\centering
\caption{Zero-shot performance of fine-tuned SAM2-T on the CAD dataset~\cite{bideau2016s}. The image encoder and the mask decoder are fine-tuned. We evaluate with box prompt.}
\label{tab:res_finetune_cad}
\resizebox{\textwidth}{!}{%
\begin{tabular}{c|ccccccc}
\toprule
~~~~~~\textbf{Method}~~~~~~ & ~~~~$S_m \uparrow$ ~~~~ & ~~~~ $F_\beta^\omega \uparrow$~~~~ & ~~~~MAE $\downarrow$~~~~ & ~~~~$F_\beta \uparrow$~~~~ & ~~~~$E_m \uparrow$~~~~ & ~~~~mDice $\uparrow$~~~~ & ~~~~mIoU $\uparrow$~~~~ \\ \midrule
SAM2-T~\cite{ravi2024sam} & 0.852 & 0.803 & 0.016 & \textbf{0.826} & 0.951 & 0.810 & 0.694 \\
SAM2-T(fine-tuned) & \textbf{0.854} & \textbf{0.807} & \textbf{0.015} & 0.825 & \textbf{0.958} & \textbf{0.817} & \textbf{0.704} \\
\bottomrule
\end{tabular}
}

\end{table*}
We fine-tune the SAM2-Tiny model on the MoCA-Mask dataset for 50 epochs. When evaluating, we use the box-based prompt of the first frame as input, the results are shown in Tab.~\ref{tab:res_finetune}. We observe notable performance improvements. The mIoU increases by approximately 0.029, from 0.623 to 0.652. Similarly, the mDice score sees a significant improvement of 0.028, from 0.728 to 0.756. The table further breaks down the improvements for different fine-tuning configurations. Fine-tuning both the image encoder and mask decoder simultaneously yields the best result. These results demonstrate the effectiveness of fine-tuning the image encoder and the mask decoder of SAM2-T on the MoCA-Mask dataset for improving segmentation accuracy in VCOS. We also evaluated our fine-tuned SAM2-T on the CAD with the box prompt, as shown in Tab.~\ref{tab:res_finetune_cad}. The CAD is evaluated in the zero-shot setting since the test cases are not seen during fine-tuning. The mIoU increases by 0.01, from 0.694 to 0.704, showing the model's domain adaptability.

\subsection{Failure Cases Analysis}
\begin{figure*}[t]
    \centering
    \includegraphics[width=\linewidth]{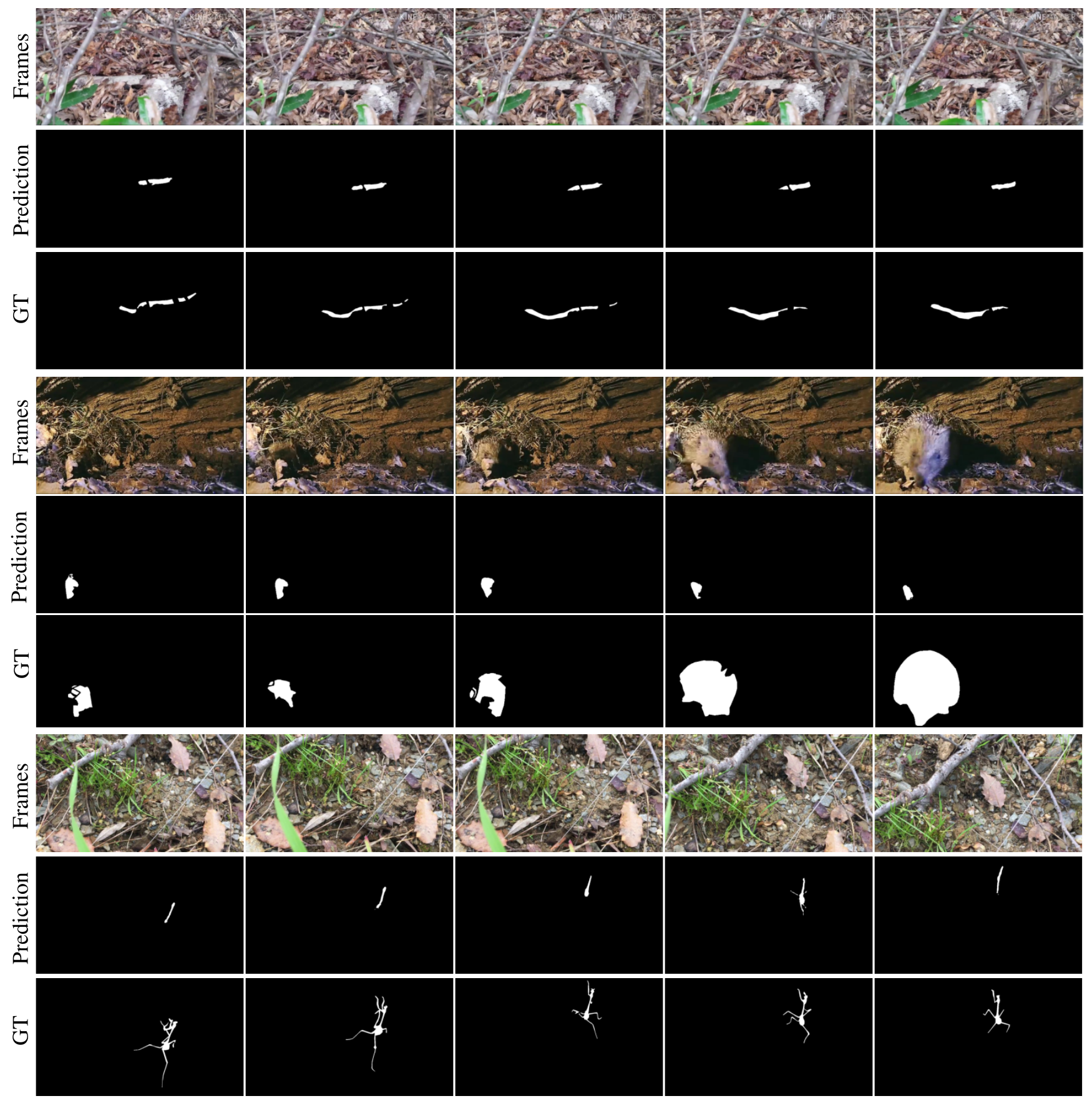}
    \caption{Failure cases of SAM2 on MoCA-Mask.}
    \label{fig:failure_case}
\end{figure*}

\begin{figure*}[t]
    \centering
    \includegraphics[width=\linewidth]{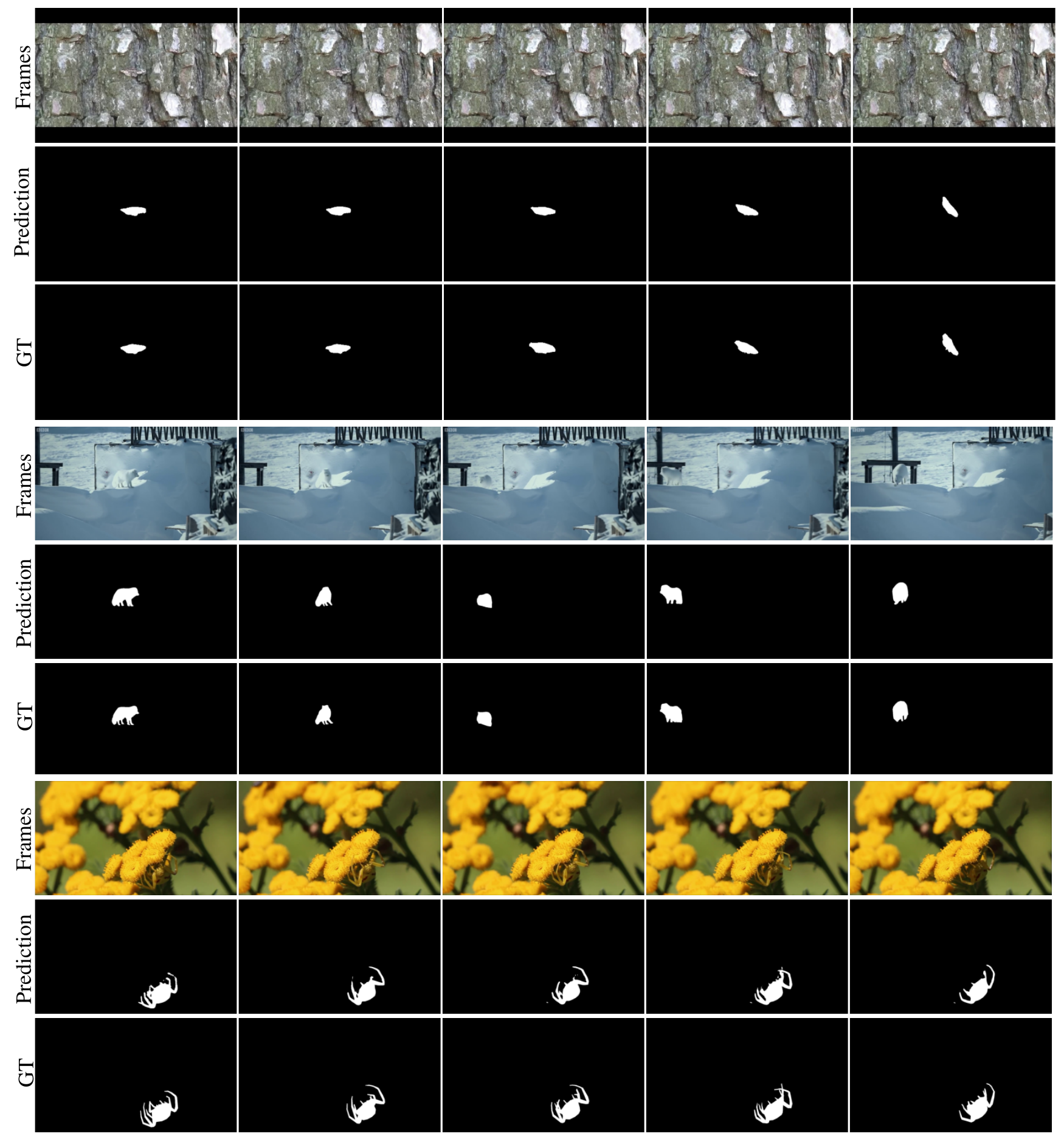}
    \caption{Qualitative examples of SAM2 on MoCA-Mask.}
    \label{fig:qualitative_case}
\end{figure*}

In our experiments, we identify several failure cases in SAM2's segmentation of camouflaged animals, using a 1-click prompt at the first frame, as shown in Fig.~\ref{fig:failure_case}. These examples, including the copperhead snake, hedgehog, and stick insect, compared with the qualitative cases (in Fig.~\ref{fig:qualitative_case}), reveal challenges when dealing with complex camouflaged scenarios. Here, we analyze the key factors that contribute to the failure:

\noindent 1) \textbf{Low Contrast with Background:}  The copperhead snake blends almost seamlessly with the surroundings, leading to an incomplete segmentation mask as SAM2 struggles to differentiate it from the background.

\noindent 2) \textbf{Occlusions and Distracting Elements:} The hedgehog case highlights SAM2's difficulty with occlusions caused by the cluttered environment, including branches and leaves, resulting in inaccurate object segmentation.

\noindent 3) \textbf{Thin and Complex Structures:} The stick insect's fine details are not well captured, especially its thin legs, showing limitations in segmenting intricate structures.

These failure cases highlight the need for improved handling of low-contrast, cluttered environments, and complex structures in future adaptations of SAM2 for VCOS.

\section{Conclusion}
This study provides a systematic evaluation of the Segment Anything Model 2 (SAM2) in video camouflaged object segmentation across two popular datasets MoCA-Mask and CAD. Our experiments highlight several key findings:

\noindent 1) \textbf{Prompt-Based Segmentation:} SAM2 demonstrates notable performance improvements with different prompting strategies, such as clicks, boxes, and masks. Box-based and mask-based prompts consistently outperformed click-based prompts across both datasets, with prompting at the middle frame often yielding the best results. This indicates the importance of spatial details (box and mask) in guiding SAM2 for accurate segmentation of camouflaged objects. 
   
\noindent 2) \textbf{Automatic Mode Performance:} In automatic mode, SAM2 struggles with fully unsupervised segmentation in camouflaged scenes, achieving unsatisfactory mIoU and mDice scores. The results show that user guidance or prompts are crucial for SAM2 to effectively segment camouflaged objects in dynamic environments.
    
\noindent 3) \textbf{Refinement with SAM2:} Using SAM2 to refine outputs from VCOS models significantly improves segmentation performance. SAM2's advanced mask refinement capabilities result in higher mIoU and mDice scores. 
    However, when combining SAM2 with Multimodal Large Language Models (MLLMs), the results were suboptimal, highlighting the importance of accurate initial object detection by MLLMs. Nevertheless, despite its limitations, this approach still outperforms some existing VCOS methods, demonstrating its potential for further refinement.
    
\noindent 4) \textbf{Fine-tuning SAM2:} Fine-tuning SAM2 on the MoCA-Mask dataset leads to clear improvements in mIoU and mDice scores, underscoring SAM2's adaptability to camouflaged segmentation when provided with specific training data. Fine-tuning SAM2 can significantly enhance its performance for tasks involving complex object-background blending, such as camouflaged segmentation.

Overall, SAM2 demonstrates strong capabilities in video camouflaged object segmentation through effective prompting strategies, model refinement, and dataset-specific fine-tuning. These findings suggest that SAM2 is a promising segmentation model for challenging camouflaged scenarios and has great potential for further improvement.

\section*{Abbreviations}
\begin{itemize}
    \item Abbreviations. COD, camouflaged object detection; CSU, camouflaged scene understanding; MLLM, multimodal large language model; SAM2, segment anything model 2; VCOS, video camouflaged object segmentation.
\end{itemize}

\section*{Declarations}
\begin{itemize}
% \item Author contributions
\item Availability of data and materials. The models and source code are released at \url{https://github.com/zhoustan/SAM2-VCOS}. The datasets used in this study are publicly available: MoCA-Mask (\url{https://xueliancheng.github.io/SLT-Net-project}), CAD (\url{https://vis-www.cs.umass.edu/motionSegmentation}).
\item Competing interests. The authors declare that they have no conflict of interest or competing interests.
\item Funding. No funding was received to assist with the preparation of this manuscript.
\item Authors' contributions. GS designed the concept and method of the work. All authors jointly helped experiments and analyze the results. YZ and GS both contributed to the implementation. YZ and GS wrote the paper while other authors carefully refined it. GX, YL, LB and EK provided valuable suggestions on how to improve the manuscript.
\item Author details. 1. Computer Vision Laboratory, ETH Z\"urich, Sternwartstrasse 7, 8092, Z\"urich, Switzerland.
2. Department of Informatics, University of Z\"urich, Binzm\"uhlestrasse 14, 8050, Z\"urich, Switzerland.
3. Integrated System Laboratory, ETH Z\"urich, Gloriastrasse 35, 8092, Z\"urich, Switzerland.
4. School of Computer Science and Engineering, Nanjing University of Science and Technology, Nanjing, China.
5. Department of Electrical, Electronic, and Information Engineering, 
University of Bologna, Via Zamboni 33, Bologna, 40126, Bologna, Italy.
\item Acknowledgments. Not Applicable.

\end{itemize}

\bibliographystyle{unsrt}
\bibliography{reference}

\end{document}